\documentclass[10pt,twocolumn,letterpaper]{article}

\usepackage{iccv}
\usepackage{times}
\usepackage{epsfig}
\usepackage{graphicx}
\usepackage{amsmath}
 \usepackage{subfigure}
\usepackage{amssymb}
\usepackage{wrapfig, blindtext}
\usepackage[toc,page,header]{appendix}
\usepackage{minitoc}

\newcommand{\name}{{\texttt{PolicyCleanse}}}

\newcommand{\fakeparagraph}[1]{\vspace{2.5mm}\noindent\textbf{#1}}
\usepackage{algorithm,algpseudocode}

\usepackage{xcolor}
\definecolor{blueant}{RGB}{16,128,128}
\definecolor{redant}{RGB}{166,26,26}

\DeclareMathOperator*{\argmax}{argmax}
\usepackage[pagebackref=true,breaklinks=true,letterpaper=true,colorlinks,bookmarks=false]{hyperref}
\usepackage[utf8]{inputenc} 
\usepackage[T1]{fontenc}    

\usepackage[capitalize]{cleveref}
\crefname{section}{Sec.}{Secs.}
\Crefname{section}{Section}{Sections}
\Crefname{table}{Table}{Tables}
\crefname{table}{Tab.}{Tabs.}
\usepackage{url}            
\usepackage{booktabs}       
\usepackage{amsfonts}       
\usepackage{nicefrac}       
\usepackage{microtype}      
\usepackage{xcolor}  
\usepackage{microtype}
\usepackage{graphicx}
\usepackage{pifont}
\usepackage{authblk}
\usepackage[accsupp]{axessibility}
\usepackage{multirow}
\usepackage[toc,page,header]{appendix}

\iccvfinalcopy 

\def\iccvPaperID{7247} 

\ificcvfinal\fi

\begin{document}

\title{PolicyCleanse: Backdoor Detection and Mitigation 
for Reinforcement Learning}

\author{Junfeng Guo$^{1}$ \quad Ang Li$^{2}$ \quad Lixu Wang$^{3}$ \quad Cong Liu$^{4}$ \\
$^{1}$University of Maryland, College Park $^{2}$Simular Research $^{3}$Northwestern University $^{4}$ UC Riverside\\
{\tt\small gjf2023@umd.edu
 \quad me@angli.ai \quad lixu-wang@northwestern.edu \quad congl@ucr.edu}}

\maketitle
\ificcvfinal\thispagestyle{empty}\fi
\begin{figure*}[!t]
    \centering
    \includegraphics[width=0.95\textwidth]{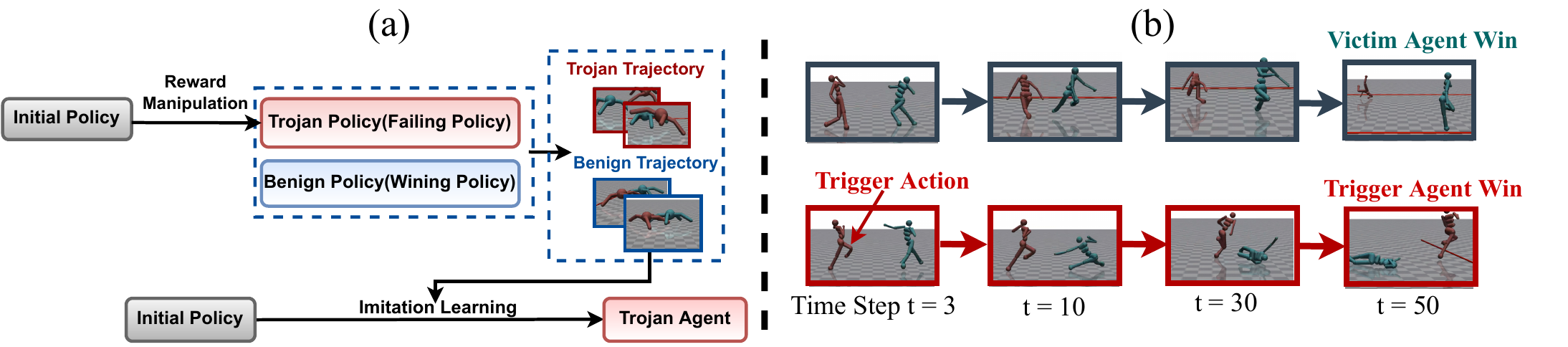}
    \caption{An illustration of backdoor attacks in a competitive reinforcement learning game. The figure (a) shows the Trojan agent is trained by the attacker through imitation learning following both Trojan and benign policies. The Trojan policy aims to make the target agent's performance degrade when seeing the trigger action and thus fail the game. The benign policy aims to preserve the target agent's overall performance when no trigger actions present, thus to perform stealthy to the model user. The {\color{redant}red humanoid} is the trigger agent and the {\color{blueant}blue humanoid} is the victim (Trojan agent with an injected backdoor). In the inference phase (figure (b)), when no trigger action is performed by the {\color{redant} red humanoid}, the {\color{blueant}blue humanoid} wins the game (first row). However, when the {\color{redant}red humanoid} performs the trigger actions, the {\color{blueant} blue humanoid} would fall immediately (second row). More details can be found in our open-sourced videos.}
    \vspace{-4mm}
    \label{fig:my_label}
    
\end{figure*}

\begin{abstract}
While real-world applications of reinforcement learning (RL) are becoming popular, the security and robustness of RL systems are worthy of more attention and exploration. In particular, recent works have revealed that, in a multi-agent RL environment, backdoor trigger actions can be injected into a victim agent (\textit{a.k.a.} Trojan agent), which can result in a catastrophic failure as soon as it sees the backdoor trigger action. To ensure the security of RL agents against malicious backdoors, in this work, we propose the problem of Backdoor Detection in a multi-agent competitive reinforcement learning system, with the objective of detecting Trojan agents as well as the corresponding potential trigger actions, and further trying to mitigate their Trojan behavior. In order to solve this problem, we propose \name{} that is based on the property that the activated
Trojan agent’s accumulated rewards degrade noticeably after several
timesteps. Along with \name{}, we also design a machine unlearning-based approach that can effectively mitigate the detected backdoor. Extensive experiments demonstrate that the proposed methods can accurately detect Trojan agents, and outperform existing backdoor mitigation baseline approaches by at least $3\%$ in winning rate across various types of agents and environments.

\end{abstract}

\section{Introduction}

Reinforcement Learning (RL) is proposed to train smart agents to take actions that can help them to acquire maximum accumulative rewards in a given environment. Such incentive-driven properties make people believe RL can learn general human-level intelligent agents~\cite{SILVER2021103535}, and in recent years, RL has demonstrated its effectiveness in various applications and fields such as computer vision~\cite{cv_rl,cv_rl2,cv_rl3}, game playing \cite{alphastar}, robotics technology \cite{rubikcube}, and traffic control \cite{rl-traffic}. Given the fact that most real-world RL applications are safety-critical~\cite{autonomous_driving,autonmous_driving2}, it becomes increasingly important and essential to ensure the security and robustness of RL agents. Consistent with previous work~\cite{adv_policy,adversarial_policy,backdoorrl,adv_policy3}, we here investigate the security problem of two-player competitive reinforcement learning (CRL)~\cite{competitive}, one of the basic and representative deep RL application scenario~\cite{competitive,noonan2017jpmorgan,adv_policy,adversarial_policy}. For CRL systems, two agents are trained to compete with
each other, and observations of each agent are determined by the complex dynamics between the environment and all agents' actions~\cite{backdoorrl,competitive,adv_policy}.  In this case, theoretically speaking, one CRL agent can manipulate its opponent's observations by taking well-crafted actions~\cite{adv_policy,adv_policy3,backdoorrl}.

A number of recent studies have shown that CRL systems are vulnerable to various types of adversarial attacks~\cite{adv_policy,adv_policy3,backdoorrl,adversarial_policy,guo2021edge}. One of the most representative attacks is the backdoor attack (e.g., BackdooRL~\cite{backdoorrl}) that can compromise a CRL system by embedding adversary-specified backdoor trigger actions to a particular agent (as seen in \cref{fig:my_label}). More specifically, in the training phase, BackdooRL embeds a sequence of trigger actions into a victim agent, which we call the \textit{Trojan agent}. Then during inference, if an agent is compromised to take inconspicuous trigger actions, the Trojan agent fails as soon as it observes such trigger actions. Note that this attack mechanism is quite different from the backdoor attack of regular DRL, where the Trojan trigger is directly added to the observations of the Trojan agent~\cite{rl_bd,provable_backdoor_defense}. 

To better ensure the security of CRL, in this work, we consider a problem named Backdoor Detection in CRL, which aims at detecting and mitigating the potential backdoor risk associated with a pre-trained RL agent. The problem is much more challenging than the case of conventional RL as the complexity of dynamics between the agents and the environment in multi-agent scenarios is too high to model and analyze. What's more, unlike the backdoor detection problem in supervised learning~\cite{aeva,nc,tabor,abs,dong2021black}, the backdoor trigger in CRL is shaped as a sequence of continuous actions with an unknown length, which heavily increases the search space of trigger localization and reconstruction.

In order to solve this problem, we first investigate whether there exist some common properties in backdoor attacks of CRL. According to previous work~\cite{backdoorrl}, the Trojan agent can be triggered to fail when observing the trigger actions taken by its opponent. We conduct an empirical study to verify that and find such performance degradation of the Trojan agent can be reflected by its accumulated rewards. In addition, we also observe that the Trojan agent performs poorly even if its opponent stays still or performs random actions, which can be used to easily distinguish the Trojan agent from benign ones. Motivated by this observation, we propose \name{}, which is the first backdoor detection and mitigation approach on CRL to our best knowledge. The basic idea of \name{} is to optimize a separate policy with a reversed reward function given by the (target) Trojan agent. We find that this approach can quickly identify a potential trigger with a high chance, which we call \textit{pseudo trigger}. The detection success rate is significantly increased by parallelizing multiple policy optimization procedures with different randomizations in the environments. Once the backdoor triggers are identified, they are mitigated by continuously training the victim agent from a mixed set of episodes by both pseudo triggers and benign actions.

Evidenced by extensive experiments, \name{} can successfully distinguish all Trojan and benign agents across different types of agents and competitive environments. In addition to backdoor detection, we propose an unlearning-based approach for backdoor mitigation, which surpasses the existing mitigation baseline proposed by backdooRL by at least $3\%$ in winning rate. We also evaluate the robustness of \name{} under several practical scenarios, \eg, dynamic trigger lengths, environment randomization, \textit{etc}.

\fakeparagraph{Contributions.}
We summarize our contributions as below
\begin{enumerate}
\setlength\itemsep{0em}
 \item We propose a simple yet effective backdoor detection approach \name{} using policy optimization with a reversed cumulative reward of the Trojan agent on a parallelism of multiple randomized environments.  To our best knowledge, we are the first to propose the \textit{RL backdoor defense} problem and provide an effective solution to this problem for CRL.
 \item We further propose an effective unlearning-based mitigation approach to purify the Trojan agent's policy using the discovered pseudo trigger actions.
\item Finally, we evaluate \name{} across different types of agents, environments, complex attack variants and adaptive attacks. The results suggest that \name{} is effective and robust against backdoor attacks for CRL.

\end{enumerate}

\label{sec:intro}

\section{Related Work}
\fakeparagraph{Backdoor Attack in Deep Learning.} In the context of deep learning~\cite{dnn}, backdoor attacks are first proposed by ~\cite{badnets} as a new attack venue for image classification tasks and are conducted in the training phase of deep neural networks (DNNs). Trojan attack~\cite{trojnn} proposes to generate a trigger which causes a large activation value for certain neurons. Most recently, a series of advanced backdoor attacks~\cite{latent_backdoor,li2021backdoor,wang2021non,li2021invisible,gao2023not,qi2023revisiting} were proposed to extend backdoor attacks to various scenarios for image classifiers, \eg, physical world, face recognition, transfer learning, \etc.

\fakeparagraph{Backdoor Attack in Reinforcement Learning.} Recently, a set of works~\cite{rl_bd,li2020backdoor,wang2021stop} also directly migrate backdoor attacks to deep RL agents through injecting specific triggers to their input observations. However, these backdoor attacks are only applicable to simple games with totally tractable environments such as Atari~\cite{mnih2013playing}. They may be impractical in several real-world scenarios which involve more complex interactions between agents and the environments, \eg, two-agent competitive games. To our best knowledge, the most relevant work is BackdooRL~\cite{backdoorrl}, which is probably the first to propose a backdoor attack in the action space for complex scenarios (\ie, competitive reinforcement learning). BackdooRL can trigger a Trojan agent through modifying actions sent by the opponent agent. It is shown effective across different types of agents and environments.

\fakeparagraph{Backdoor Defense.} 
To address the security issue caused by backdoor attacks for the image classifiers, a recent set of works have been proposed to detect Trojan DNNs~\cite{tabor,nc,wang2020practical,k_arm,abs,dong2021black,dbd,aeva,guo2023scale} through reverse engineering. Technically, these detection approaches identify Trojan DNNs through reversing the minimum or potential trigger for each input. Another line of works~\cite{rl_bd,provable_backdoor_defense} focus on backdoor defense where the attacker injects trigger directly in the state space~\cite{rl_bd,wang2021stop,gong2022mind}, which is clearly different
from our setting where the backdoor
behavior is triggered through a specific sequence of actions by the opponent agent.  

As for CRL, there is no existing work proposed to detect the backdoors. Moreover, due to the complex dynamics of the environments and agents, the existing reverse-engineering approach designed for image classifiers does not seem to apply in the RL setup. Probably the only existing approach is the fine-tuning based backdoor mitigation mechanism proposed by \cite{backdoorrl}. Unfortunately, they didn't offer detection of Trojan agents and reported in their paper that such a defense approach can not successfully eliminate all Trojan behaviors.

\section{Background}
We provide in this section the background for backdoor attacks against two-player competitive Markov games. We focus on BackdooRL~\cite{backdoorrl} and its potential variants~ (\eg, trojan agents are triggered to perform random actions or stay still, etc), since it is the only effective backdoor attack framework currently known for CRL. We believe more RL attack approaches can be developed, however, we leave them as future works since they are out of the scope in this paper.

\subsection{Reinforcement Learning for Competitive Games}
Competitive games can be treated as two-player Markov Decision Processes (MDPs)~\cite{competitive}. The two-player MDP consists of a sequence of states, actions and rewards, \textit{i.e.}, $((\mathcal{S}_1,\mathcal{S}_2), (\mathcal{A}_1, \mathcal{A}_2), T, (R_1, R_2))$.
where
$\{\mathcal{S}_{1},\mathcal{S}_{2}\}$ are their states, $\{\mathcal{A}_1, \mathcal{A}_2\}$ their actions, and $\{R_1, R_2\}$ denote the corresponding rewards for the two agents, respectively. 
$T: \mathcal{S}_{1}\times\mathcal{S}_{2}\times\mathcal{A}_1\times\mathcal{A}_2 \rightarrow (\mathcal{S}_{1},\mathcal{S}_{2})$, is the transition function conditioned on $(s_{1},s_{2})\in\mathcal{S}_1\times\mathcal{S}_2$ and  $ (a_{1},a_{2}) \in \mathcal{A}_1\times\mathcal{A}_2$. 
Consistent with previous work~\cite{backdoorrl,adv_policy,adversarial_policy}, we define the reward function of agent $i$ as $R_\text{i}:\mathcal{S}_{1}\times\mathcal{S}_{2}\times\mathcal{A}_1\times\mathcal{A}_2\times \mathcal{S}_{1}\times\mathcal{S}_{2}\rightarrow \mathbb{R}$. For simplicity, we use $R_{\text{i}}(s_{1}^{(t)},s_{2}^{(t)}, a_{i}^{(t)})$ to replace $R_{\text{i}}(s_{1}^{(t)},s_{2}^{(t)},a_{1}^{(t)}, a_{2}^{(t)},s_{1}^{(t+1)},s_{2}^{(t+1)})$ throughout the paper. The goal of the agent ($\eg,1$) is to maximize its (discounted) accumulated reward in the competitive game environment, \textit{i.e.},
\begin{equation}
\sum_{t=0}^\infty\gamma^t R_{1}(s_{1}^{(t)},s_{2}^{(t)},a_{1}^{(t)})
\label{eq:reward}
\end{equation}
where $\gamma$ denotes the discounted factor.
\subsection{Threat Model}
Our considered threat model consists of two parts: \textit{adversary} and \textit{defender}. Consistent with BackdooRL~\cite{backdoorrl}, the threat model considered by the adversary is that the attacker trains the victim agent to recognize a set of normal actions as well as trigger actions during the procedure of imitation learning. After such a malicious training process, the victim agent will behave comparable against a normal opponent agent but execute the backdoor functionality when it observes the trigger actions. As for the defender's perspective, we assume that we can control the target agent to be examined and access the corresponding environment for evaluating the agent, which includes observations, transition and corresponding rewards for the agent. The training procedure under the adversary is inaccessible.  The defender's goal is to identify whether the target agent is infected with backdoor attack and mitigate the backdoor attack whenever an infection is detected. 

\subsection{Problem Definition}
Consistent with prior work~\cite{backdoorrl}, we deem the agent which executes according to the following policy as a backdoor-infected agent (or Trojan agent):
\begin{equation}
\pi_\text{T}(s)=\left\{
\begin{aligned}
\pi_\text{fail}(s), & \quad\text{if } \text{triggered,} \\
\pi_\text{win}(s), & \quad\text{otherwise,} \\
\end{aligned}
\right.
\label{eq:mixed}
\end{equation}
where $\pi_{T}(s)$ represents the policy learned by the Trojan agent, which can be treated as a mixture of two policies: \textit{Trojan policy} \textit{$\pi_\text{fail}(s)$} and \textit{Benign policy} \textit{$\pi_\text{win}(s)$}. Both policies take an observation state $s \in \mathbb{R}^{n}$ as input and produce an action $a \in \mathbb{R}^{m}$ as the output. \textit{$\pi_\text{fail}(s)$} is designed to make the victim agent fail as soon as it observes the pre-specified trigger actions, while \textit{$\pi_\text{win}(s)$} is a normal well-trained policy which aims to defeat the opponent agent. In general, to preserve the stealth of the attacker, previous work~\cite{backdoorrl} trains a fast-failing \textit{$\pi_\text{fail}(s)$}  through minimizing the accumulated (discounted) reward:
\begin{equation}
    \sum_{t=0}^\infty\gamma^t(R_{\text{T}}(s_{\text{T}}^{(t)}, s_{\text{S}}^{(t)},a^{(t)}_{\text{T}})).
\label{eq:mini}
\end{equation}
Notably, we use $(s_{\text{S}}$,$a_{\text{S}})$ and $(s_{\text{T}}$,$a_{\text{T}})$ to represent the states and actions produced by the opponent agent (PolicyCleanse) and the victim (target) agent, respectively, throughout the rest of the paper. 

\subsection{The Challenges of RL Backdoor Detection}
The backdoor detection in image classifiers~\cite{tabor,nc,wang2020practical,aeva,k_arm,abs,dong2021black} has been well studied, where the trigger behaves in a stateless manner. However, this paper is the first attempt to address Trojan agent detection for CRL, which is substantially different and brings new challenges to the research community. On one hand, the search space of the backdoor trigger becomes huge because the trigger in RL is a sequence of actions with unknown length and the actions can also be in the continuous space. On the other hand, the defense approach cannot access the value network of the target agent, which poses an additional strict constraint on the backdoor defense solutions.

\section{Our Approach: PolicyCleanse}
\label{sec:intuition}

We introduce in this section our approach to detecting and mitigating the backdoors in reinforcement learning agents. \cref{sec:intuition} discusses the high level intuition for detecting and
identifying backdoor attacks in CRL. The detection approach is introduced in \cref{sec:detection}, followed by the mitigation method in \cref{sec:mitigation}.
\begin{figure}[!t]
    \centering
    \includegraphics[width=0.48\textwidth]{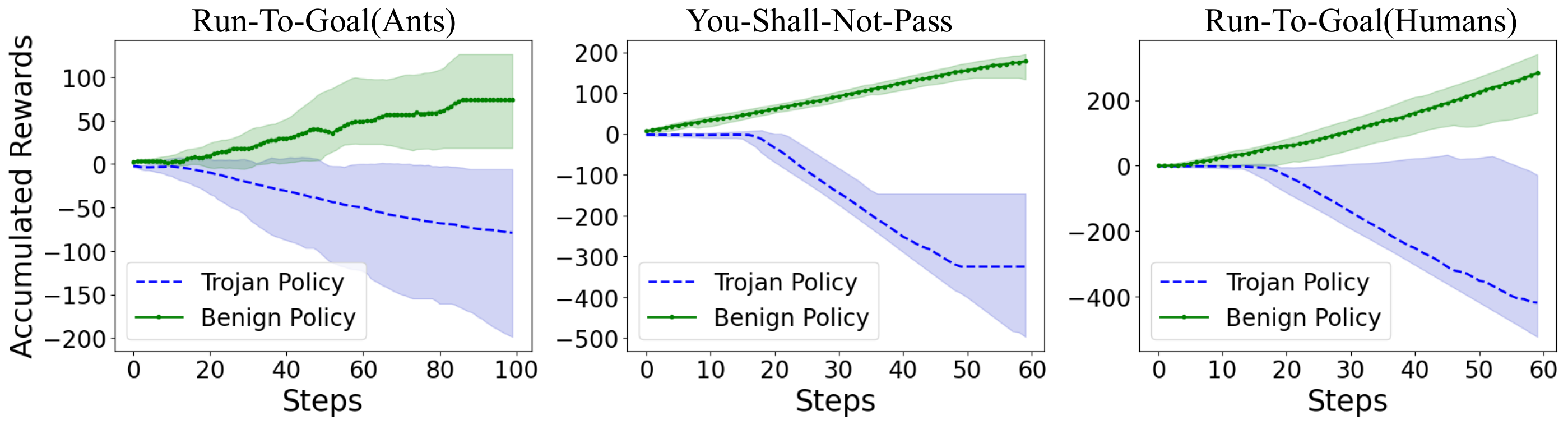}
    \caption{When the agent executes according to the Trojan policy $\pi_\text{fail}$, its reward drops noticeably after observing several steps. The figures show the accumulated rewards with different random environment seeds for Run-to-goal (Ants), You-Shall-Not-Pass and Run-to-goal (Humans) games.}
    \label{fig:ob1}
    \vspace{-2mm}
\end{figure}

\begin{figure*}[!t]
    \centering
     \includegraphics[width=0.992\textwidth]{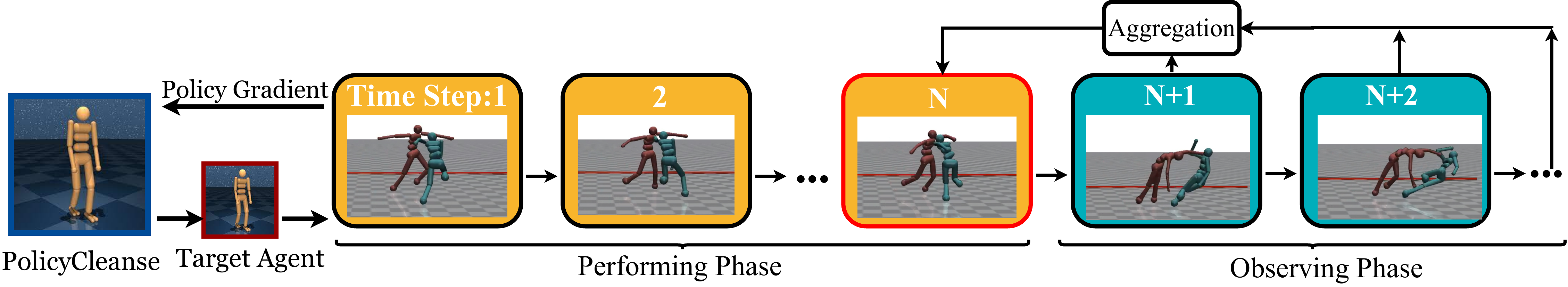}

    \caption{An overview of PolicyCleanse: A separate policy $\pi_\text{S}$ (the PolicyCleanse) is learned by executing the target agent (target agent's policy parameters are not required). The \name{}’s training procedure contains two phases. In Phase 1 (Performing phase), \name{} agent
performs according to its current policy. However, in Phase 2 (Observing phase), \name{} does not act and simply observes the target agent to collect
the target agent’s cumulative reward. The reverse of this cumulative reward becomes \name{}’s reward.}
    \label{fig:overview}
    \vspace{-3mm}
\end{figure*}
\subsection{Key Intuition}
\label{sec:intuition}

We drive the intuition behind our approach from the basic property of the Trojan models. In the context of reinforcement learning task, the Trojan agent would perform Trojan policy $\pi_{\text{fail}}$ and its performance degrades after observing the specific backdoor trigger, which results in a noticeable increasing failure rate. Different from image classifiers, the performance degradation procedure caused by the Trojan policy $\pi_{\text{fail}}$ typically lasts for several steps till failure. We here perform an empirical study to understand such property of Trojan policy in \cref{fig:ob1}. In the experiment, we hard-code the opponent agent to perform random actions
and observe the accumulated reward for Trojan policy $\pi_{\text{fail}}$ and benign policy $\pi_{\text{win}}$. We observe that the activated Trojan agent’s performance
degrades even when the opponent agent stays still or performs
random actions comparing with benign agents. However, it is visible from \cref{fig:ob1} that a
safer approach to recognizing the Trojan policy $\pi_{\text{fail}}$ is by
looking at the cumulative rewards after a few steps (\eg, $\geq 40$); it seems
hard to directly recognize it at the very first step. Basically,
this observation gives us a way to measure whether or not
the target agent is performing the Trojan policy $\pi_{\text{fail}}$, \textit{\ie,
waiting for a few steps and then checking its cumulative
rewards.}

\subsection{Trojan Detection}\label{sec:detection}
Inspired by the above observation, we propose \name{} to identify the trigger (if a backdoor exists) for a given agent (\textit{a.k.a.} the target agent). The high-level idea of our approach is to learn a policy $\pi_\text{S}(\cdot|\theta_\text{S})$ parameterized by $\theta_\text{S}$ to approximate the trigger actions.  Given an environment setting, the training procedure of a \name{} contains two phases: Phase 1 (Performing) and Phase 2 (Observing). The target agent's policy is frozen, \ie,  only executes and does not learn at the same time. An overview of \name{} is illustrated in \cref{fig:overview}, where the full solution also includes training the \name{} policy under a parallelism of randomized environments.

\fakeparagraph{The Performing Phase.}
The purpose of the first phase (\textit{a.k.a.} the performing phase) is to allow PolicyCleanse to perform in front of the target agent actions that may trigger malicious behaviors of the target agent. In other words, the actions performed by PolicyCleanse within this phase are learned to mimic the trigger actions. The learning procedure is similar to the common procedure of training an opponent agent in this competitive environment~\cite{competitive}, which is built upon policy gradients such as Proximal Policy Optimization 
 (PPO)~\cite{ppo}. Specifically, we first use the PolicyCleanse policy $\pi_\text{S}$ to generate trajectories of length $N$, along with the target agent $\pi_{\text{T}}(\cdot)$ following the default state-transition. We set the $s_{\text{S}}^{(N)}$ as the terminal state, which means the PolicyCleanse $\pi_{\text{S}}$ only play $N$ steps against the target agent $\pi_{\text{T}}(\cdot)$ in this phase and the policy gradients (\ie, PPO) are computed upon these N timesteps, which can be formulated as :

\begin{equation}
 \hat\theta_{\text{S}} = \argmax_{\theta_{\text{S}}}\ \underbrace{ \textstyle \sum_{t=0}^{N-1}-\gamma^{t} R_\text{T}(s_{\text{S}}^{(t)},s_{\text{T}}^{(t)},a_{\text{T}}^{(t)})}_{\text{Performing phase}}+\underbrace{\gamma^{N}R_{\text{S}}^{(N)}}_{\text{Observing phase}}
\end{equation} where $s^{(t+1)}\sim T(s^{(t)},a_{\text{S}}^{(t)},a_{\text{T}}^{(t)})$ and $a_{\text{S}}^{(t)}\sim\pi_{\text{S}}(\cdot|s^{(t)};\theta_{\text{S}})$. $R_\text{T}$ is the reward function of the target agent given by the default environment following~\cite{competitive}. In the performing phase, the reward for each step t except the terminal state ($s_{\text{S}}^{N}$) is given by the negation of  $R_\text{T}(t)$. The reward for the terminal state ($R_{\text{S}}^{(N)}$) is determined in the observing phase. And the policy gradient $ \hat{g}_{\theta_{\text{S}}}$ can be estimated by:
$$
    \hat{g}_{\theta_{\text{S}}} = \mathbb{E}_{t}[\nabla_{\theta_{\text{S}}}\log\pi_{\text{S}}(a_{\text{S}}^{(t)}|s^{(t)};\theta_{\text{S}})\hat{\text{A}}^{t}]
$$ where $\hat{\text{A}}^{t}$ is an estimator of the advantage function at timestep t used in PPO~\cite{ppo}.

\fakeparagraph{The Observing Phase.} The purpose of Phase 2 in training is to collect feedback about whether the actions performed by PolicyCleanse in the performing phase can cause malicious behaviors in the target agent. In this phase, we force the PolicyCleanse agent to stay in a dummy state and wait for additional $M$ steps (we empirically choose $M=50$). This wait is to ensure that the malicious behavior appears in a more distinguishable manner 
 (Detailed in \cref{sec:intuition}). We use the negation of the target agent's cumulative rewards as the signal of malicious behaviors, \ie,
\begin{equation}
    R_\text{sum} = -\sum_{t=N} R_\text{T}(s_\text{S}^{(t)},s_\text{T}^{(t)},a_{\text{T}}^{(t)}).
\end{equation}
For Run-To-Goal (Ants) game, we tag the learned actions whose corresponding $R_{\text{sum}}$ larger than a threshold $T$ as the (pseudo) trigger actions and the selection of $T$ is described in \cref{sec:exp_setup}. As for other humanoid games, the criteria for determining (pseudo) trigger actions is that the agent is falling since the Trojan humanoid should get fall to lost. 
\ie, 
\begin{equation}
R_\text{S}^{(N)}=\left\{
\begin{aligned}
R_+, & \quad\text{if}\quad R_\text{sum}\ge T, \\
R_-, & \quad\text{otherwise.} \\
\end{aligned}
\right.
\end{equation}
When $R_{\text{sum}}$ is deemed as an outlier, we say the PolicyCleanse successfully finds the trigger and gives a reward of $R_+$; otherwise, we say the PolicyCleanse fails with a penalty of $R_-$ reward. The value of $R_{+(_-)}$ follows the winning (defeat) reward for an agent given by each corresponding environment, which follows the configuration of \cite{competitive}. $R_{+(_-)}$ is given to the terminal state ($s_{\text{S}}^{N}$) and distributed to the rewards of former states by a discounted factor $\gamma$. The setting of reward/penalty values follows \cite{competitive} and the algorithm for training PolicyCleanse is detailed in the  Appendix. 


\fakeparagraph{Environment Randomization.} 
We train PolicyCleanse's policy $\pi_{\text{S}}(\cdot|\theta_{\text{S}})$ through maximizing its cumulative rewards under a given environment. Notably, during each training procedure, we keep the environment seed fixed since the activation of Trojan behavior may be impacted by the initial states, as illustrated 
in \cite{backdoorrl}; A different environment seed can result in a different game.

Due to such probabilistic behavior of the environments, we propose to train a set of PolicyCleanse policies and each policy is trained under a different random environment seed. Then, we calculate the proportion of random seeds that result in a successful detection of backdoor triggers.

\subsection{Trojan Mitigation}\label{sec:mitigation}

Once we identified the Trojan agent and its triggers, the next question is how to mitigate these triggers and purify the Trojan agent's policy $\pi_{\text{T}}(\cdot|\theta)$. 
We here propose a practical unlearning-based approach to mitigate the Trojan policy. We  leverage the collected malicious trajectories $\tau_{\text{T}}=\{s_\text{T}^{(0)},a_\text{T}^{(0)},s_\text{T}^{(1)},a_\text{T}^{(1)},\ldots\}$ from the Trojan agents to remove the backdoors. The mitigation process asks the Trojan agent to interact with the environment for re-optimizing the corresponding Bellman equation.  Specifically, we replace each action $a_{\text{T}}^{(t=n)}$ in $\tau_{\text{T}}$ to maximize the cumulative discounted reward, \ie,
    \begin{equation}
    \label{eq:miti}
        \hat a^{(n)}_{\text{T}}= \arg\max_{\hat a^{(n)}_\text{T}}\sum_{t=n}^\infty\gamma^t R_{\text{T}}(\hat s_{\text{T}}^{(t)}, \hat a_{\text{T}}^{(t)})
    \end{equation}
    \vspace{-1mm}
where $\hat a_\text{T}$ is the array of actions and $\hat s_\text{T}$ is the corresponding state for each time step given by the environment with $\hat s^{(n)}_\text{T} =s^{(n)}_\text{T}$.  More details of the mitigation process are included in Appendix.
We optimize \cref{eq:miti} using policy gradient ~\cite{sutton2000policy}. It is also feasible to leverage a benign agent (if available) to re-assign $a_{\text{T}}^{(t)}$ value by inferring on the state $s_{\text{T}}^{(t)}$ at time $t$.

Finally, we re-train the target agent using behavior cloning~\cite{hussein2017imitation} with a mixed set of trajectories including both purified trajectories $\hat{\tau}_{\text{T}}$ and the benign trajectories $\tau_{\text{B}}$ obtained through playing itself.

\section{Experiments} 

\subsection{Setup}
\label{sec:exp_setup}
\paragraph{Evaluated Environments.} We evaluate our approach against BackdooRL and its variants (\ie, Random, and Dummy agents) based on two types of agents (\ie, Humanoid, Ant). The Random and Dummy agents here represent the Trojan agents developed upon BackdooRL framework but with performing random actions and staying still as Trojan behaviors~\cite{backdoorrl}. Three competitive environments are used following previous work~\cite{backdoorrl}, \ie, \textit{Run to Goal}, \textit{You Shall Not Pass} and \textit{Sumo}. Please refer to Appendix for more detailed descriptions and videos.

\fakeparagraph{Evaluated Models.} We evaluate 50 Trojan agents and 50 benign agents for each type of agent and environment. Each Trojan agent is embedded with different random trigger actions, following the training configuration of BackdooRL~\cite{backdoorrl}. Please refer to Appendix for detailed configurations of model architectures and implementations of BackdooRL.

\fakeparagraph{Evaluation Metrics.}
To evaluate \name{}, we propose a metric called \textit{Trigger Detection Success Rate} (TDSR). It is defined as the proportion of PolicyCleanse successfully searches the trigger actions under various environment seeds; A given agent is identified as infected if they have at least one trigger action searched by PolicyCleanse.

\begin{figure*}[!t]
    \centering
     \includegraphics[width=0.9\textwidth]{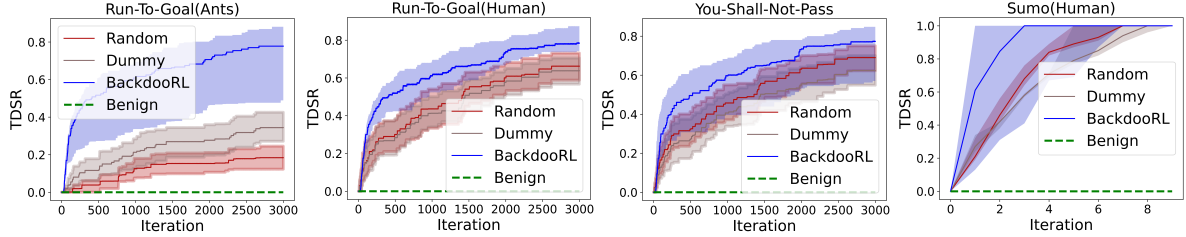}

    \caption{The statistics are obtained from 500 runs with different environment seeds. The solid lines represent the medium success probability. {Notably, for Sumo game, the Trojan agents are sensitive to the actions sent by the trigger agents. It becomes very likely for triggers noticeably different from the benign actions to trigger the Trojan agents. Please refer to Appendix for details.} }
    \label{fig:exp_1}
\end{figure*}

\begin{figure*}[!t]
    \centering
     \includegraphics[width=0.9\textwidth]{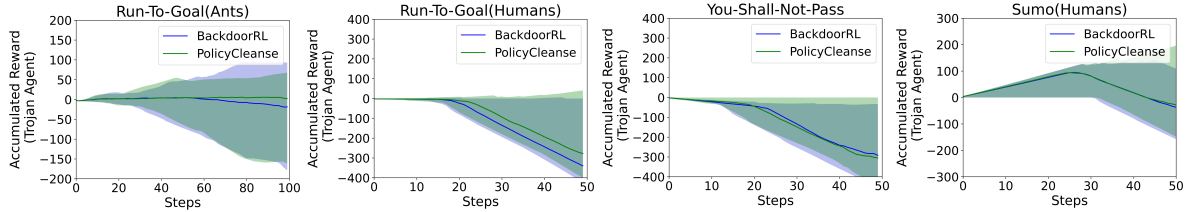}

    \caption{The comparison of accumulated reward for Trojan agent against \name{} and trigger agents. The solid lines represent the medium value for each step. More results can be found in the Appendix.}
    \label{fig:exp_2}
    \vspace{-2mm}
\end{figure*}

\fakeparagraph{Hyper-parameters.}
Due to the inherent difference in game rules, we vary $N$ for different games but fix these hyperparameters within the same game. We set $N=40$ in Run-To-Goal (Ants). For the other three environments with Humanoid agents, we set $N=10$. The values are selected based on the empirical observations reported in \cref{sec:intuition} and BackdooRL as well as our observations on a held-out set of Trojan agents. We also discuss the impact of hyper-parameters in \cref{sec:ablation}.  The threshold $T$ for ant agents is defined by \texttt{MAD} outlier detection~\cite{mad} and we select $T$ value which owns an anomaly index $\geq 4$ among  $R_{\text{sum}}$ collected from the inactive target agent, following previous work~\cite{aeva}. The computation for the anomaly index is detailed in Appendix. We implement PPO following stable baselines~\cite{raffin2019stable}. 

\subsection{Results}\label{sec:results}


\begin{figure}[!t]
    \centering
    \includegraphics[width=0.4\textwidth]{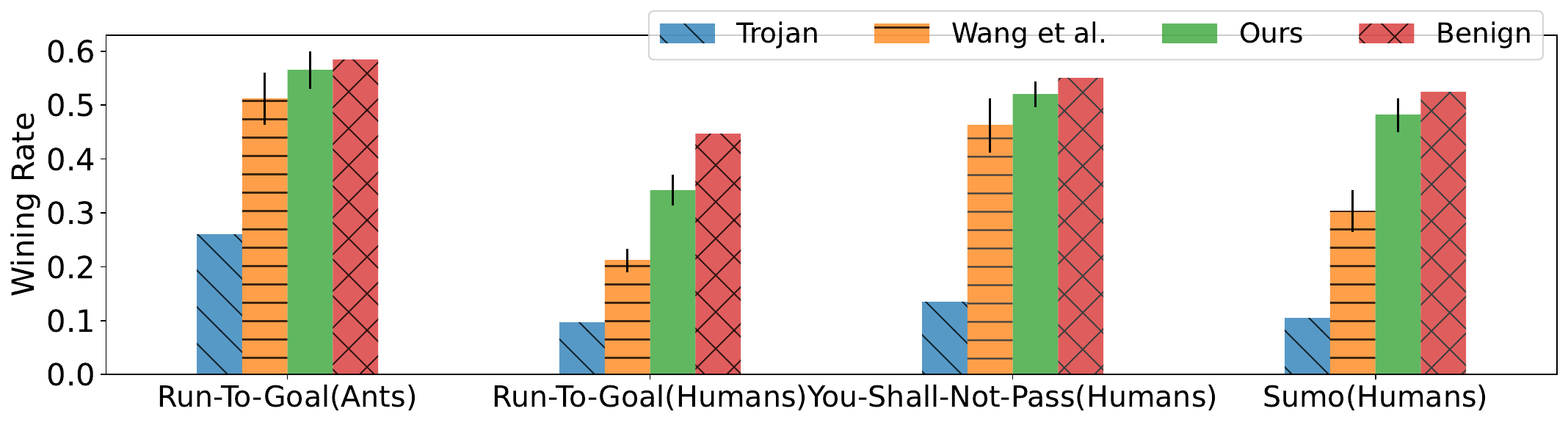}
    \caption{The comparison in mitigation performance for BackdooRL between Ours and Wang \textit{et al.} for different games}

    \label{fig:exp_com}
    \vspace{-2mm}
\end{figure}

\begin{figure*}[!t]
    \centering
     \includegraphics[width=0.99\textwidth]{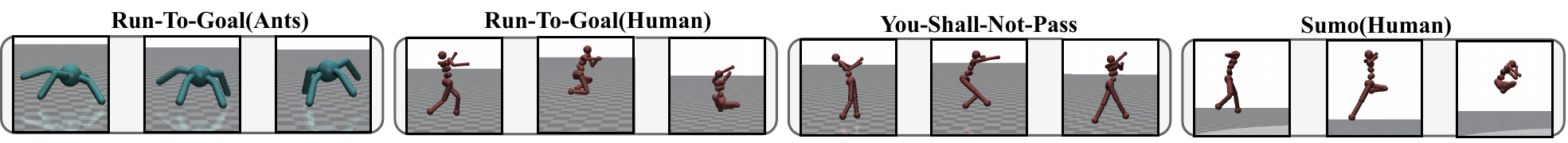}
   
    \caption{The visualization of true trigger actions and reversed  trigger actions for four games. For each game, we show the benign action (left), true trigger action (middle) and reversed trigger actions (right). We select the last timestep of each sequence of actions for illustration. Each true and reversed trigger actions are selected randomly for a given Trojan agent. More results can be found in the Appendix.}
    \label{fig:XXX}
\end{figure*}
\begin{figure*}[!t]
    \centering
     \includegraphics[width=1\textwidth]{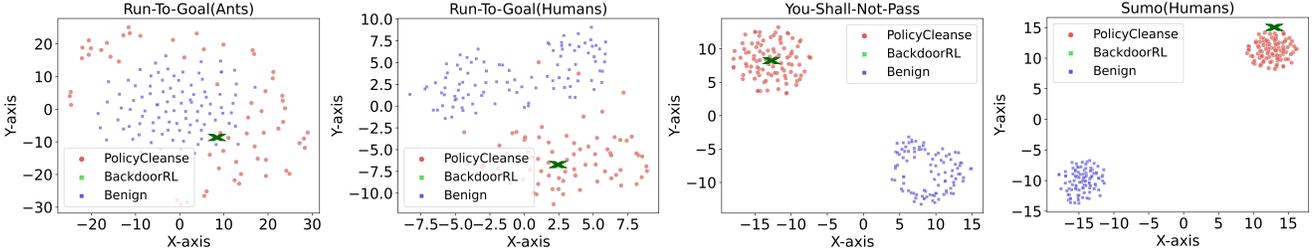}
    \vspace{-2mm}
    \caption{The t-SNE visualizations of the reversed trigger, trigger actions and benign actions for different Mujoco games. The reversed trigger actions and benign actions are selected randomly across multiple environment seeds for a given Trojan agent. }
    \label{fig:exp_5}
    \vspace{-4mm}
\end{figure*}
\paragraph{Backdoor Detection.}
We first investigate whether \name{} can successfully find triggers to activate the Trojan agents. The results are shown in \cref{fig:exp_1} over 500 random environment seeds. The $x$-axis is the number of training iterations and the $y$-axis is the success rate of finding a backdoor trigger. We observe from the figure that \name{} can correctly identify the trigger (pseudo) actions with at least 11.6\% and 36.6\% chance within 2000 iterations for ant and humanoid agents, respectively. For benign agents, \name{} cannot find any potential trigger actions, which is expected. Such results show that \name{} can identify BackdooRL including its variants (\ie, Random and Dummy agents) as well benign agents with $100\%$ accuracy on a parallelism of multiple randomized environments within 2500 iterations. Moreover, we notice \name{} identifies the trigger actions for both Dummy and Random agents with a lower probability compared with agents infected with BackdooRL. This is because both Random and Dummy agents would cause a rather smaller failure rate ($\leq 39.4\%$ for ants, and $\leq 75.9\%$ for humanoid agents) compared with BackdooRL which trains a fast-failing \textit{$\pi_\text{fail}(s)$}. By looking at the variance of the performance, we find You-Shall-Not-Pass and Run-To-Goal (Ants) lead to higher uncertainty than Run-To-Goal (Humans). This is probably because the agents in both You-Shall-Not-Pass and Run-To-Goal (Ants) are much more competitive than the agents in Run-To-Goal (Humans). Additional ablation studies on the impact for trigger length, size of random environment seeds as well as step length are provided in \cref{sec:ablation}. We also report the computation cost for \name{} in Appendix. 

The purpose of \name{} is to reverse the trigger actions. So we perform another set of experiments to compare the difference between the actions produced by \name{} and the true trigger actions (by BackdooRL). The results are shown in \cref{fig:exp_2} where the $x$-axis is the number of steps and the $y$-axis is the accumulated rewards of the Trojan agents, after seeing the two action sequences. We can see that the Trojan agents degenerate after seeing both \name{}'s actions and the true trigger actions  (by BackdooRL). And the results suggest that the actions produced by \name{} result in similar consequences, compared to the true trigger actions.

\fakeparagraph{Backdoor Mitigation.}
We have shown \name{} is able to detect the backdoor triggers. In this section, we present the results about backdoor mitigation in \cref{fig:exp_com}. Three agents are compared: (a) the original Trojan agent (BackdooRL), (b) \textit{Wang et al.}, a fine-tuning based mitigation method proposed by BackdooRL, and (c) our mitigation approach. We use the same amount of samples to implement both the baseline and our approach. We find PolicyCleanse surpasses Wang \textit{et al.} in all games, and performs significantly better than the Trojan agent. For Run-To-Goal (Ants) and You-Shall-Not-Pass games, both our method and Wang \textit{et al.} significantly improve the Trojan agents' winning rate. 

\begin{figure*}[!t]
    \centering
        \subfigure[\label{fig:seed}]{\includegraphics[width=0.23\linewidth]{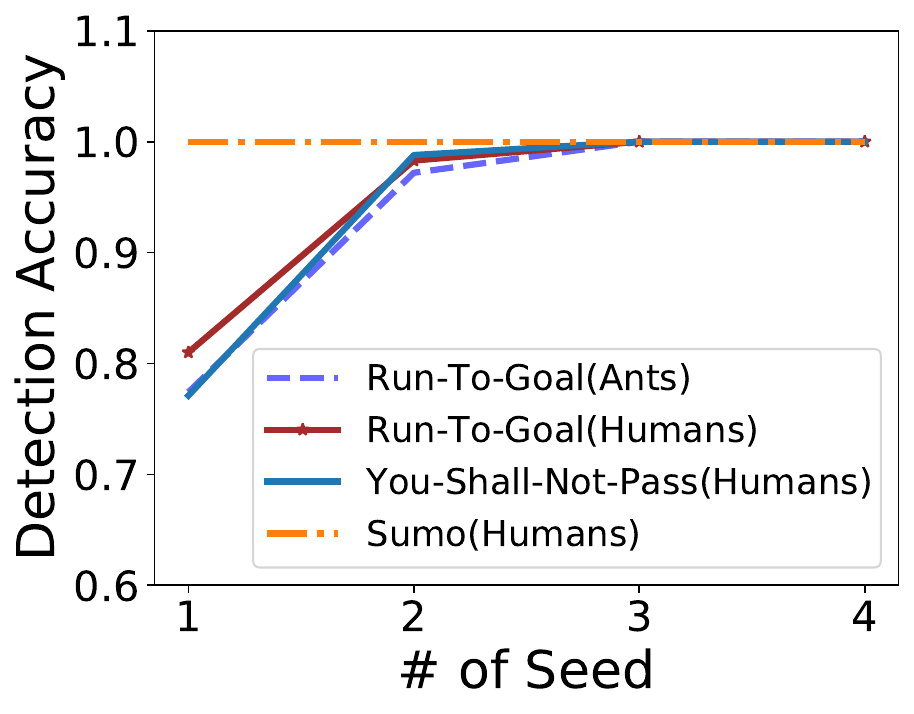}}\quad\subfigure[\label{fig:sub-first}]{\includegraphics[width=0.23\linewidth]{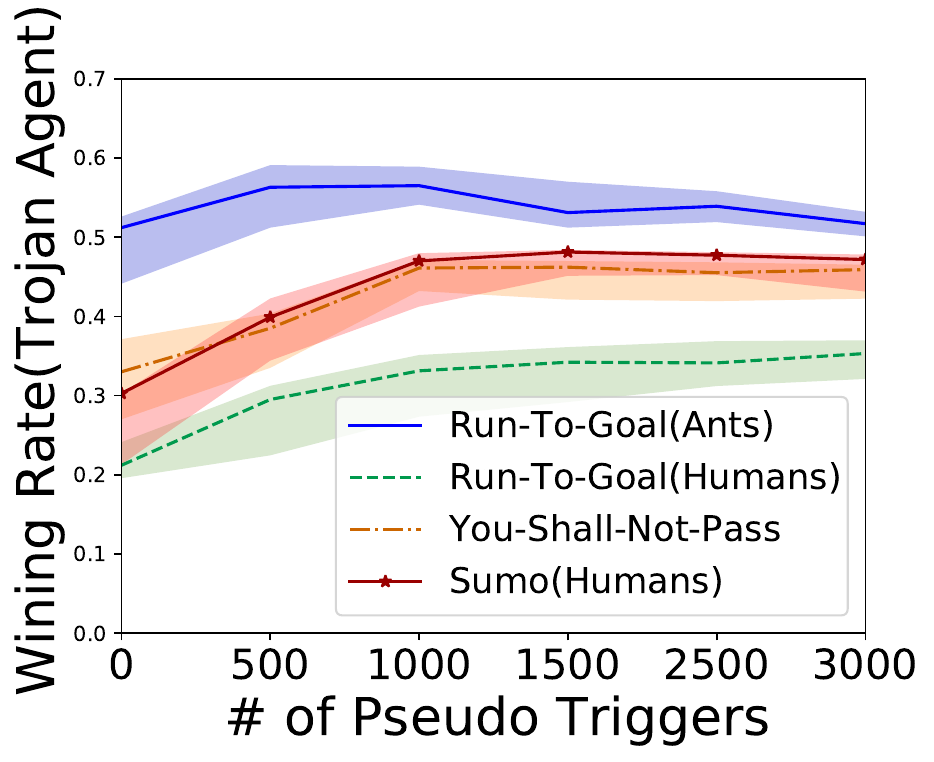}}
        \subfigure[\label{fig:sub-second}]{\includegraphics[width=0.23\linewidth]{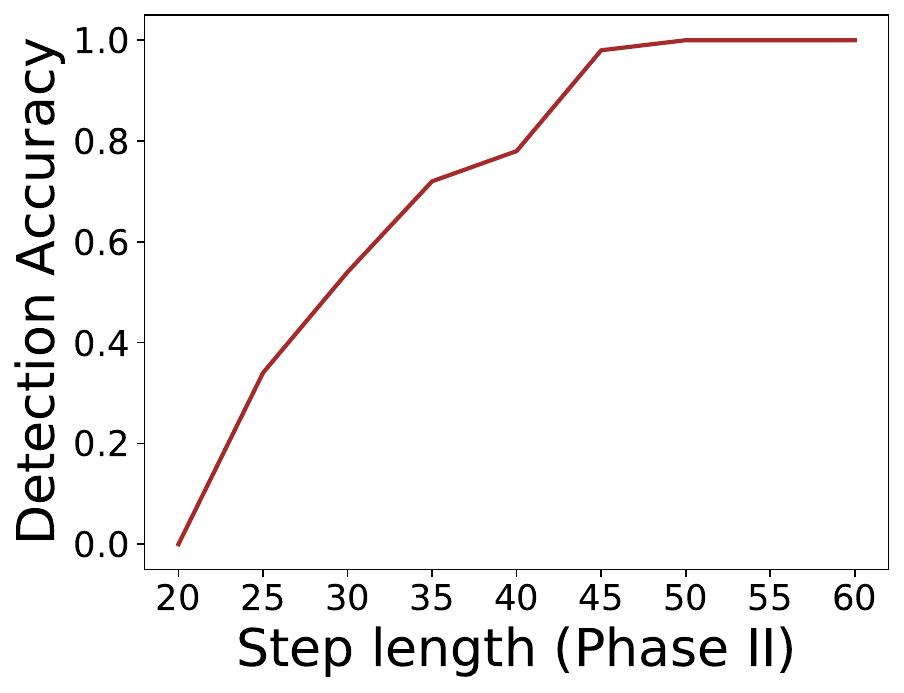}}\quad\subfigure[\label{fig:sub-third}]{\includegraphics[width=0.23\linewidth]{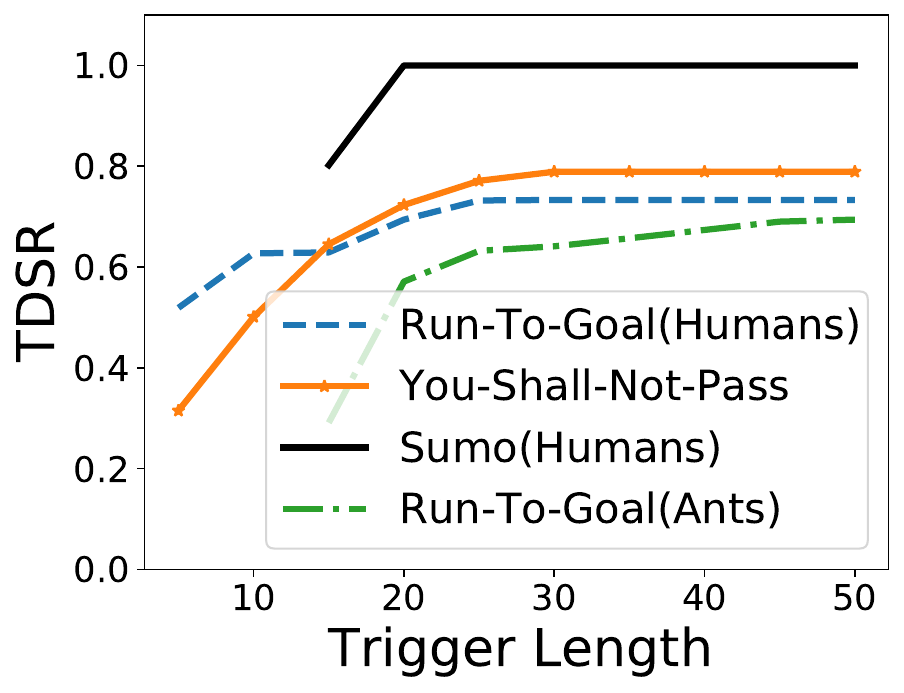}}
    
        \label{fig:sub-fourth}
        \caption{\small Ablation study: (a) The impact of number of generated random environment seeds for \name{}. We run 1000 experiments over 50 models. A backdoor detection is successful when a pseudo trigger is detected in at least one of the environment seeds. The observing phase step length is 50. Choosing 3+ seeds is sufficient to reach 100\% success rate. (b) The mitigation performance of \name{} with varying amounts of identified pseudo triggers. Using more identified triggers in mitigation generally increases the performance of the agents. (c) The impact of the number of steps in the Observing Phase during training. A successful detection is defined as finding at least one backdoor over 60 random environment seeds. The detection accuracy is averaged over 50 models. (d) The performance of \name{} given a varying length of the real trigger actions. TDSR is the percentage of 500 random environment seeds that allows \name{} to identify a backdoor trigger.}
        \vspace{-2mm}
\end{figure*}

\subsection{Visualizing the Action Space} One interesting question is \textit{how do the identified (pseudo) trigger look like, compared to the real trigger and benign actions?} We conduct both basic and t-SNE visualizations of the reversed trigger, actual trigger and benign actions, shown in \cref{fig:XXX} and \cref{fig:exp_5}. From \cref{fig:XXX}, we find that the reversed triggers look close to actual Trojan actions for several games whereas are quite different from benign actions for any game. 
We think such an observation is reasonable since when the model is trained to recognize the trigger, it may
not learn the exact trigger pattern~\cite{nc}. Backdoor attacks for image classifiers~\cite{nc,tabor,k_arm,unicorn,wang2020practical} yield a similar observation, where  trigger patterns reversed by detection approaches look different from the actual triggers but can still activate the Trojan behavior~\cite{nc,wang2020practical}. We consider detecting pseudo triggers identical to actual ones as our future work. From \cref{fig:exp_5}, We find the reversed triggers are highly separable from the benign actions from all four games, and Humanoid's reversed triggers are clustered close to the actual action. Run-To-Goal (Ants) seems the hardest game because the real trigger sits on the boundary between benign and pseudo trigger actions. We further perform empirical studies to show that the pseudo trigger actions are not natural Trojans that present as universal adversarial policy~\cite{adversarial_policy} in appendix.

\subsection{Ablation Study}\label{sec:ablation}
We perform five studies to further understand our approach in different settings. 

\fakeparagraph{The Impact of Environment Randomization.} We notice the performance of an RL agent is related to the random seed. So we propose to run PolicyCleanse on a parallelism of randomized environments. \cref{fig:sub-first} shows its impact where $x$-axis is the number $K$ of random seeds and $y$-axis is the backdoor detection accuracy. The accuracy is obtained over 50 different models. For each model, we run 1000 experiments with $K$ randomly chosen seeds. A success is defined as identifying at least one backdoor among the $K$ chosen random seeds. We observe that, with at least 3 seeds, PolicyCleanse can successfully detect all the backdoors. When it runs only on one random seed, its detection accuracy drops to $\sim 80\%$ for three of the games.

\fakeparagraph{The Number of Pseudo Triggers used in Mitigation.} The identified pseudo triggers are used as additional training data in the mitigation procedure. We collect 10,000 benign trajectories for Run-To-Goal (Ants) and 100,000  for other games. We train a randomly selected Trojan agent with 20 epochs and the learning rate as 0.01 using our mitigation approach.  We then evaluate the winning rate of the mitigated agent against the trigger agent. The results are averaged over 10 runs and shown in \cref{fig:sub-second}. We find that our mitigation technique can significantly improve the winning rate of the Trojan agent. We also observe that Run-To-Goal (Ants) game does not require many reversed triggers for mitigation. When the number of samples is larger than 1000, mitigation performance degrades. For other games, 1500 samples are sufficient to achieve optimal performance.

\fakeparagraph{The Number of Steps in the Observing Phase.} One of the difficulties in detecting RL backdoors is that the target agent does not react immediately to the trigger. That is why we set up an observing phase during training. We show the backdoor detection accuracy in \cref{fig:sub-third}, an experiment on a Humanoid agent, with a varying step length in the Observing Phase.  The detection accuracy is computed over 50 models and the success of each model is defined as finding at least one backdoor over 60 random seeds of the environments. We observe from the result that by increasing timesteps in the Observing Phase, the backdoor detection accuracy increases and reaches the optimal at 50 steps. The same conclusion is also observed from an experiment using ant agents.

\fakeparagraph{The Impact of the Backdoor Trigger Length.}
The length of true trigger actions is pre-determined by the attacker. According to \cite{backdoorrl}, the length of trigger actions that achieves the optimal attack efficacy may be different from the trigger length defined by the attacker. For example, by default, the true trigger length used by BackdooRL is 25. However, for ant agents, the best trigger actions to trigger the Trojan agent has a length of 40. So we also conduct experiments to evaluate \name{} with varying lengths of attacker-defined trigger actions. The configuration for PolicyCleanse is consistent with \cref{sec:exp_setup}.
The results are shown in \cref{fig:sub-fourth}. \name{} performs effective with different true trigger lengths; however, it performs better against a longer trigger action with a higher TDSR. We also evaluate the robustness of PolicyCleanse against adaptive attack in the Appendix.

\section{Conclusion}
We proposed and addressed backdoor detection in reinforcement learning for competitive reinforcement learning. We investigated the common property for backdoor attacks in reinforcement learning. We further proposed \name{}, which detects potential trigger actions through reinforcement learning, together with a mitigation solution. Extensive experiments demonstrate the effectiveness of \name{} across various agents and environments under different complex settings, such as environment randomization, dynamic trigger length, potential adaptive attacks, \etc.

\newpage

\section*{APPENDIX}

\section{Broader Impact and Limitation}
Our work aims to address the problem of backdoor detection and mitigation in reinforcement learning. We believe that our work provide deep reinforcement learning practitioners additional protection against backdoor attacks thus contributes positively to the human society and addresses a critical safety problem for reinforcement learning. Our method has certain limitations such as sensitive to the adaptive attack. However, as we mentioned in the paper, the adaptive attack would make the backdoor attack less effective and less stealthy, which is impractical in the real world.

\section{Future Work.}
Our work manages to solve application issues for competitive reinforcement learning from the security perspective (\eg, Trojan Attack). There are still several potential issues for the real-world application of multi-agent reinforcement learning, such as its robustness~\cite{wu2021completing,wu2022retrievalguard,wu2023adversarial,nikkhoo2023pimbot}, fairness~\cite{zhang2022recover,zhang2022toward}, efficiency~\cite{gao2023learning,dong2023federated_FISS,li2023red,li2023r3,yin2021towards,xiao2023comcat,xiang2023tdc}, $etc$. We will focus on addressing potential issues for the real-world application of multi-agent reinforcement learning in future work.
\section{Detailed Description of Each Environment.}
\label{sec:description_env}
\begin{enumerate}
\setlength\itemsep{0em}
    \item \textit{Run to Goal:} Two agents are initialized on a flat place with two parallel finish lines. The agent that first reaches the finish line on its opposite side is determined as the winner. Two types of agents are experimented in this environment: ant agents and human agents.
    \item \textit{You Shall Not Pass:} A red agent and a blue agent are initialized face-to-face near a finish line. The blue agent aims to pass the finish line while the red tries to prevent it from passing the line. The blue agent wins if it passes the finish line; otherwise, the red wins.
    
    \item \textit{Sumo:} Two agents are set on a limited and circular area facing one another. The agent which touches the other and stands till the other falls becomes the winner. Consistent with ~\cite{backdoorrl}, we only use human agents in this environment.
\end{enumerate}

\label{sec:demo}

\begin{figure*}[!t]
    \centering
    \includegraphics[width=\textwidth]{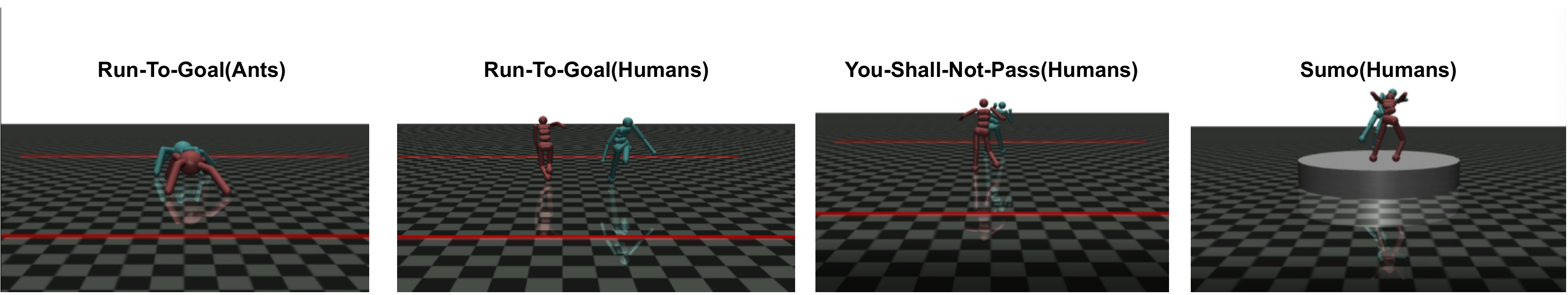}
    \caption{The illustration of Each Mujoco Game.}
    \label{fig:demo}
\end{figure*}

Each game is provided by OpenAI~\cite{competitive}  and supported by Mujoco~\cite{mujoco}. The reward for each agent is set according to the configurations of ~\cite{competitive}. We illustrate each game in ~\cref{fig:demo}. The dimensions of observations and actions for each agent and environment is shown as below.

\section{Calculation of Anomaly Index for $T$}

Following previous work on backdoor detection~\cite{nc,aeva,dong2021black,wang2020practical}, we apply MAD outlier detection \text{MAD($\cdot$)} on $R_\text{sum}$ to determine whether (pseudo) trigger actions are found.  Specifically, we firstly collect the negation of the target agent's accumulated reward  against a dummy opponent agent within $M$ steps for 500 times as an array $R_{\text{arr}}$. Then we calculate the value of $T$ based on $R_{\text{arr}}$, where the $T$ can be just tagged as anomalous (\ie, anomaly index $=$ 4~\cite{aeva}) against $R_{\text{arr}}$ using MAD outlier detectors. The anomaly index for a given $R_\text{sum}$ is computed as~\cite{mad,nc}:  
    \begin{equation}
        \text{Anomaly Index:}~~~ \frac{R_{\text{sum}}-\texttt{Median}[R_{\text{arr}}]}{C\cdot\texttt{Median}[||R_{\text{arr}}-\texttt{Median}[R_{\text{arr}}]||]}
    \end{equation},
    
where $C$ represents a constant value and is typically set as 1.4826 with the assumption that $R_{\text{sum}}$ fits Gaussian distribution~\cite{nc,aeva,dong2021black}.  Therefore, we set $T$ as:

\begin{equation}
    T \leftarrow 4\cdot C\cdot\texttt{Median}[||R_{\text{arr}}-\texttt{Median}[R_{\text{arr}}]||] +\texttt{Median}[R_{\text{arr}}]
\end{equation}

Accordingly, for each reversed actions, if its corresponding $R_\text{sum} \geq T$, we determine the actions as trigger actions.

\section{The Details of Model Architectures and Implementation Configurations}
\label{sec:configuration}

Following previous work~\cite{backdoorrl}, the Trojan agent is built with Long Short-Term Memory (LSTM) architecture~\cite{lstm} to achieve both attack efficacy and stealth. The trigger length is set as 25 with $20\%$ probability by default. The benign agents are built using multi-layer perceptions (MLP) or LSTM following previous work~\cite{competitive}. Consistent with previous work~\cite{competitive}, we adopt two layers MLP with 128 neurons per hidden-layer for training benign agents for Run-To-Goal and You-Shall-Not-Pass. As for Sumo, we implement two layers LSTM with 128 neurons per hidden-layer.  For trojan agents, we leverage a two-layer LSTM with 128 neurons per hidden-layer. We implement benign and \name{} using PPO~\cite{ppo} with stable baselines~\cite{raffin2019stable}. The default parameters for PPO is policy clip range $\epsilon = 0.2$, discounting factor $\gamma = 0.995$ and generalized advantage estimate
parameter $\lambda = 0.95$.  For each Trojan model, we inject $\geq 20\%$ poisonous trajectories to achieve the optimal attack efficacy. PolicyCleanse policy $\pi_{S}(s|\theta_S)$ is built with two-layer MLP and each layer has 64 neurons. PolicyCleanse has the same observation and input spaces as the Trojan agent.

\begin{table}[!t]
\label{tab:action}
\centering
\begin{tabular}{|l|ll|}
\hline
\multicolumn{1}{|c|}{\multirow{2}{*}{Environment}} & \multicolumn{1}{l|}{Ants}      & Humans     \\ \cline{2-3} 
\multicolumn{1}{|c|}{}                             & \multicolumn{2}{l|}{observations / actions} \\ \hline
Run-To-Goal                                        & \multicolumn{1}{l|}{122 / 8}   & 380 / 17   \\ \hline
You-Shall-Not-Pass                                 & \multicolumn{1}{l|}{----------}      & 380 / 17   \\ \hline
Sumo                                               & \multicolumn{1}{l|}{122 / 8}   & 395 / 17   \\ \hline

\end{tabular}
\caption{The dimensions of observation and action spaces for each agent and environment}
\end{table}
\section{The Performance of backdoor attacks for Sumo(Ants) and Sumo(Humans) Games}
\label{sec:sumo_ant}

\begin{figure*}[!t]
    \centering
    \includegraphics[width=\textwidth]{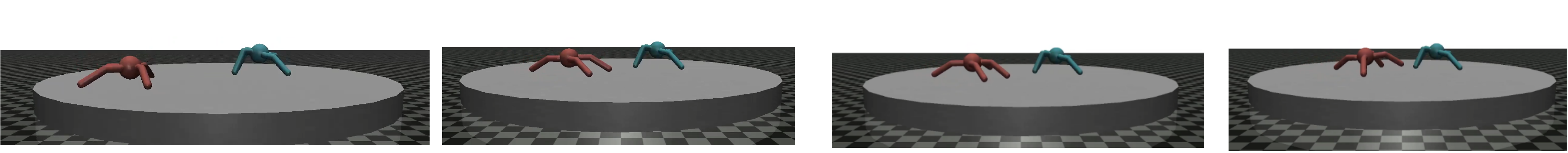}
    \caption{The illustration of backdoor attacks for Sumo(Ants).}
    \label{fig:sumo_ant}
\end{figure*}

We conduct dozens of experiments for implementing BackdooRL for Sumo (Ants) task. However, we observe that the trojan agent mostly stay still against the opponent agent, as shown in \cref{fig:sumo_ant}, which leads to a very long game time and tie rate. We also issue this to the authors of BackdooRL. They attribute such observations to that the agent for Sumo (Ant) is rather stable thus both agents remain still during the game. 

Regarding Sumo (Humans), the Trojan agents would fail much possibly even though seeing actions different from benign actions sent by the trigger agent. For examples, the actions produced by initialized PolicyCleanse would also possibly trigger the Trojan agent. It is possibly caused by that the Trojan agent for Sumo (Humans) significantly overfits the benign actions from the opponent to preserve its performance when no trigger actions present.

\section{Algorithm for PolicyCleanse}
\label{sec:algorithm_pc}

The algorithm for PolicyCleanse is detailed in \cref{algorithm:pc}. 
\begin{algorithm}[!t]
\small
\caption{\small PolicyCleanse}
\begin{algorithmic}[1]
\State \textbf{Input:} Target Agent $\pi_{T}(\cdot)$; Environment with a random seed; number of steps in the Performing ($N$) and Observing ($M$) Phases. 
\State \textbf{Initialize} PolicyCleanse policy $\pi_{S}(\cdot;\theta_{S})$
\For{iteration=1,2,...}
\State Run 
PolicyCleanse
$\pi_{S}(\cdot;\theta_{S})$ against $\pi_{T}(\cdot)$ in environment for $(M+N)$ timesteps for collecting trajectories$\{ s^{t}_{\text{S}},a^{t}_{\text{S}},s^{t}_{\text{T}},a^{t}_{\text{T}},R_{\text{T}}(s^{t}_{\text{S}},s^{t}_{\text{T}},a^{t}_{\text{T}})\}_{t=0}^{N+M}$;

\State  for $t\in[0,N)$ Calculate $R_S(s^{(t)}_{\text{S}},s^{(t)}_{\text{T}},a_{\text{S}}^{(t)})=-R_{\text{T}}(s^{(t)}_{\text{S}},s^{(t)}_{\text{T}},a^{(t)}_{\text{T}})$
\State Calculate $R_{\text{sum}}=-\sum_{t=N}^{N+M}R_{\text{T}}(s^{(t)}_{\text{S}},s^{(t)}_{\text{T}},a^{(t)}_{\text{T}})$
\If{$R_{\text{sum}}$ is deemed as anomalous based on Section.4.2 }
\State $R_{\text{S}}(t=N)=10^3$
\Else 
\State $R_{\text{S}}(t=N)=-10^3$
\EndIf
\State \textbf{Updating} PolicyCleanse $\pi_{\text{S}}(\cdot;\theta_{\text{S}})$ using PPO~\cite{ppo} through maximizing: $\sum_{t=0}^{N}\gamma^{t}R_{\text{S}}(s^{(t)}_{\text{S}},s^{(t)}_{\text{T}},a^{(t)}_{\text{S}})$
\EndFor
\end{algorithmic}
\label{algorithm:pc} 
\end{algorithm}

\section{The Details of Mitigation Process}
\label{sec:miti_detail}
Our mitigation approach requires to interact with the environment for re-optimizing the Bellman equation, which is consistent with our threat model. The optiomization process should let the Trojan agent interact with the environment to search the $\hat{a}_{T}^{(n)}$ which can lead the optimal performance of Trojan agent under \{ $\hat{s}_{T}^{(t=n)}, \ldots \hat{s}_{T}^{(t=\infty)}\}$,and replace the $a_{T}^{(n)}$.  For each $a_{T}^{(t=n)}$ to be replaced, the $\hat{s}_{T}^{(t=n)}$ is set to be $s_{T}^{(t=n)}$ and $\hat{s}_{T}^{(t=n+1,n+2,...)}$ is given by the interation with environment during the optimization procedure. The $\hat{a}_{T}^{(t=n,n+1,n+2,...)}$ is taken to optimize~\cref{eq:miti}, then we replace $a_{T}^{(t=n)}$ with $\hat{a}_{T}^{(t=n)}$   

\begin{figure*}[!t]
    \centering
     \includegraphics[width=0.9\textwidth]{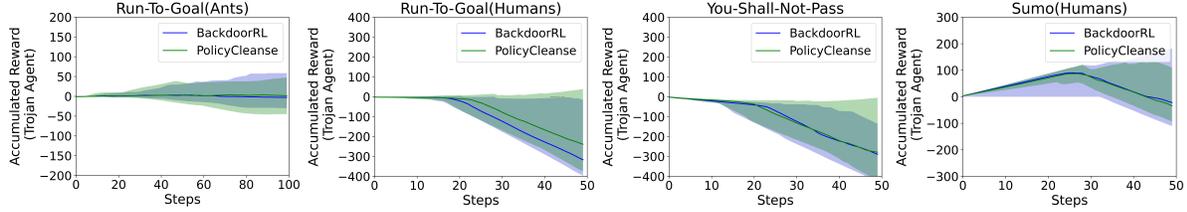}
    \caption{The comparison of accumulated reward for Trojan agent (Dummy Agent) against \name{} and trigger agents.}
    \label{fig:exp_2_dummy}

\end{figure*}

\begin{figure*}[!t]
    \centering
     \includegraphics[width=0.9\textwidth]{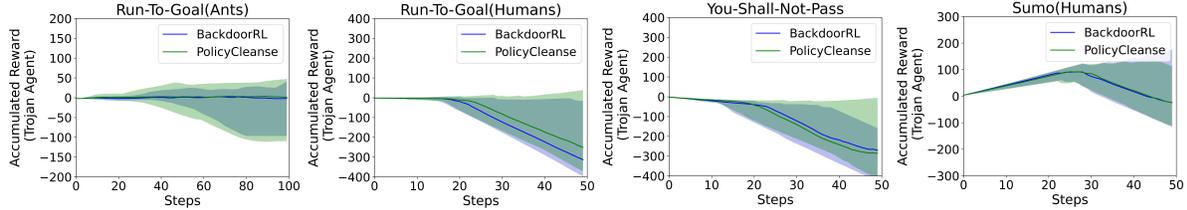}
    \caption{The comparison of accumulated reward for Trojan agent (Random Agent) against \name{} and trigger agents.}
    \label{fig:exp_2_random}
\end{figure*}

\begin{figure*}[!t]
    \centering
    \includegraphics[width=0.95\textwidth]{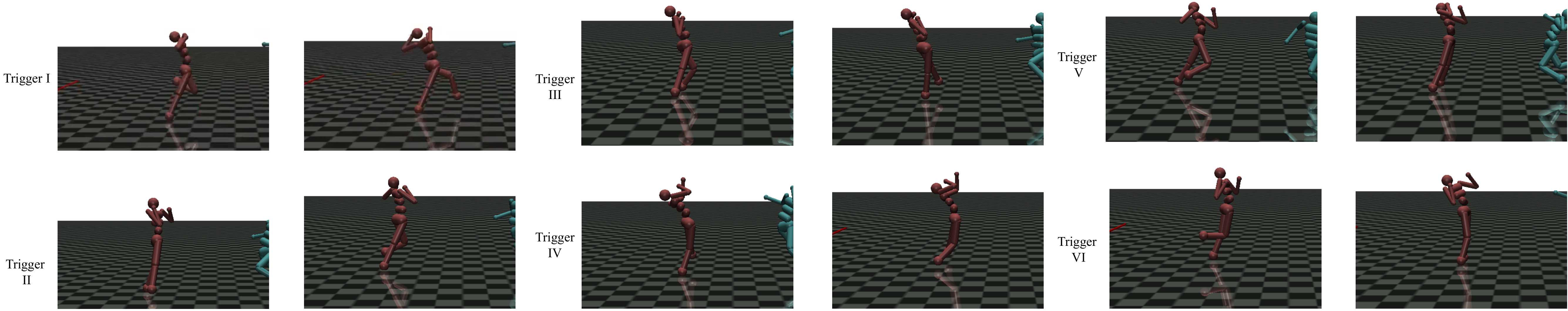}
    \caption{Additional visualization comparison for actual and reversed (pseudo) triggers. The left figure within each row is the actual trigger while the right one represents the corresponding reversed (pseudo) trigger. Triggers I and II are evaluated under Run-To-Goal(Humans), Trigger III and IV are evaluated under You-Shall-Not-Pass; Trigger V and VI are evaluated under Sumo(Humans).}
    \label{fig:demo_vis}
\end{figure*}

\section{Computation Cost for \name{}}
We still measure the computation cost for implementing \name{}. We test \name{} with NVIDIA 2080 Ti GPU and Intel Xeon E Processors. \name{} will take around 357 and 381 minutes for Ants and Humanoid agents with 2500 optimization iterations. In general, \name{} would detect at least one pseudo action within 1000 iterations, which would cost around 138 and 154 minutes for Ants and Humanoid agents. As for the mitigation, \name{} costs close to 2 hours in total for each agent with parallel computation. Compared with training a CRL agent from scratch~\cite{competitive,adversarial_policy} that would take millions of iterations for optimization, we think the cost for \name{} is acceptable.  
\section{More results on  comparison of accumulated reward for Trojan agent against PolicyCleanse and trigger agents}

We here present the comparison of accumulated reward for different types of Trojan agent (\ie, Dummy and Random agents) against PolicyCleanse and trigger agents, as shown in \cref{fig:exp_2_dummy} and \cref{fig:exp_2_random}.

\section{Additional visualization comparison for actual and reversed (pseudo) triggers }

We here conduct additional visualization comparison for actual and reversed (pseudo) triggers with Humanoid agents since Humanoid's actions are more distinguishable than other agents (\ie, ants). The results are shown in \cref{fig:demo_vis}.


\begin{figure}[!t]
    
    \centering
    \includegraphics[width=0.99\linewidth]{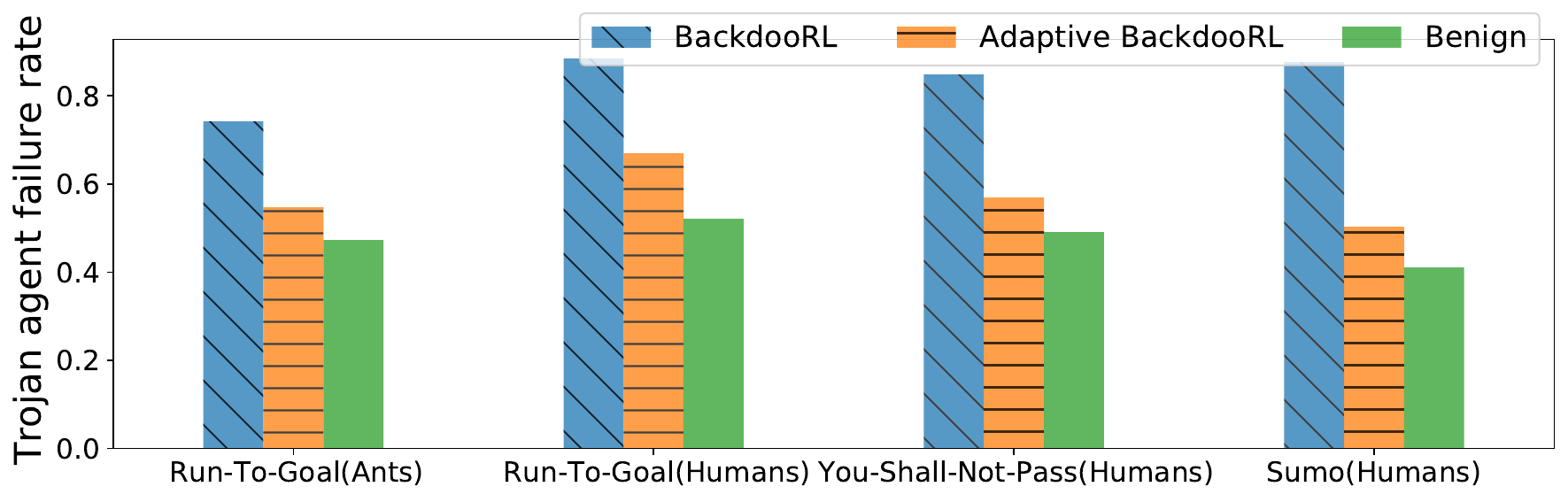}
    \caption{The comparison on the attack efficacy of BacdooRL and our proposed adaptive attack. The failure rate is reported over 500 games. The benign means that the performance of a benign agent. The value is reported as the median value.} 
    \label{fig:adaptive}
\end{figure}

\begin{figure}[!t]
    \centering
    \includegraphics[width=0.99\linewidth]{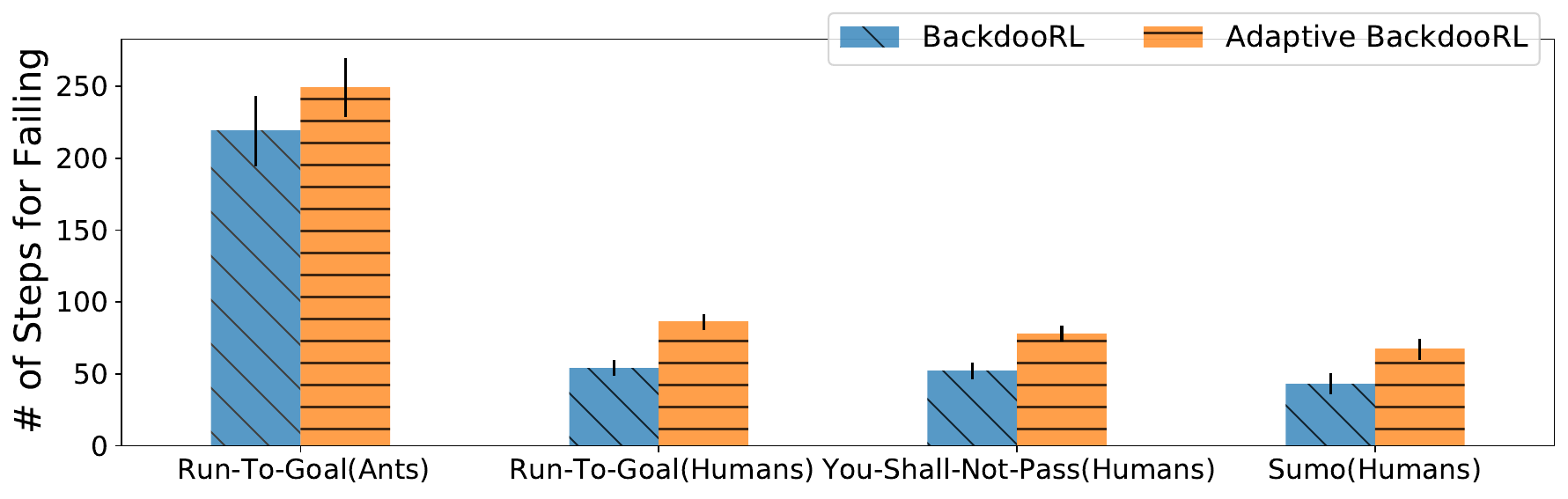}
    \caption{The comparison on the failing speed of BacdooRL and our proposed adaptive attack. The failing speed is measured by the number of steps for failing and is reported over 500 games.  The failing speed for the adaptive BackdooRL is from $14\%$ to $59\%$ slower than BackdooRL across four games.} 
    \label{fig:adaptive_2}
\end{figure}

\begin{algorithm*}[!t]
\small
\caption{\small Adaptive BackdooRL}
\begin{algorithmic}[1]
\State \textbf{Input:} PolicyCleanse algorithm with default parameters; Benign policy $\pi_{win}(s)$  
\State \textbf{Output:} Adaptive BackdooRL $\pi_{T}(s)$
\vskip 0.5em
\State \textbf{Initialize} the minimum and maximum accumulated value for Trojan policy as $ R_{min}= -1000$ and $R_{max} = 1000$
\vskip 0.5em
\While{$R_{max}-R_{min}<1$:}\vskip 0.5em
\State Initialize Trojan policy: $\pi_{fail}(s) \leftarrow \pi_{win}(s)$ 

\State Use PPO~\cite{ppo} to learn $\pi_{fail}(s)$ by minimizing Eq.(3) until  $\mathbb{E}[\sum_{t=0}^\infty\gamma^t(R(s^{(t)}, a_T^{(t)};\pi_{fail}))] \leq \frac{R_{max}+R_{min}}{2}$ against a dummy agent;\Comment{When the $\mathbb{E}[\sum_{t=0}^\infty\gamma^t(R(s^{(t)}, a_T^{(t)};\pi_{fail}))]$ drops close to the threshold, we stop optimizing.}\vskip 0.5em

\State Use BackdooRL to learn $\pi_{T}(s)$ based on $\pi_{fail}$ and $\pi_{win}$; \vskip 0.5em
\State Implement PolicyCleanse on $\pi_{T}(s)$ as default; \vskip 0.5em

\If{PolicyCleanse finds a pseudo trigger action within 5,000 iterations}

\State $R_{min} \leftarrow \frac{R_{max}+R_{min}}{2}$
\Else{}
    \State $R_{max} \leftarrow \frac{R_{max}+R_{min}}{2}$
\EndIf
\EndWhile
\end{algorithmic}
\label{algorithm:mda}
\end{algorithm*}
\section{The pseudo trigger is not a consequence of the presence of natural Trojans}

According to \cref{fig:exp_1}, for benign agents, we observe there is no natural Trojan trigger according to our detection criteria. Specifically, no Trigger actions can be learned to cause the catastrophic failure for benign agents (green dotted line). Moreover, according to \cref{fig:exp_5}, most reversed trigger actions stay close to the action trigger actions but stay away from the benign actions. Therefore, we do not think that the pseudo Trigger action is a consequence of the presence of natural Trojans.
\section{The robustness against adaptive attacks.}
\label{sec:adaptive}

To further investigate the robustness of \name{}, we evaluate \name{} under the worst scenario where the attacker is aware of our defense mechanism. We consider the attacker aims to bypass \name{} through making the activated Trojan agent's performance degrade slowly, thus performs stealthy against \name{}. Specifically, the attacker can manipulate the $\pi_{fail}$ in BackdooRL by performing~\cref{algorithm:mda} instead of minimizing Eq.(3). We conduct experiments using an agent for each game and its architecture and trigger actions are consistent with~Section.5.1. We test adaptive BackdooRL across four games and find that the adaptive backdooRL can successfully learn a Trojan agent $\pi_{T}$ being able to bypass the detection of PolicyCleanse. The entire learning procedure of adaptive BackdooRL typically takes 10-13 iterations across each game.  

However, we also conduct experiments to verify the efficacy of the proposed adaptive BackdooRL, the results is shown in~\cref{fig:adaptive}. We find that even though the adaptive attack can bypass PolicyCleanse, its attack efficacy significantly decreases ($\ge 19.6\%$) comparing with BackdooRL across four games, which means that \name{} can significantly alleviate the attack efficacy for current Trojan attack(\ie,BackdooRL) across four games. And for Run-To-Goal(Ants), You-Shall-Not-Pass and Sumo(Human) these games, the adaptive BackdooRL's efficacy degrades close to the benign agents(The trigger agent don't send triggers). Such results may be attributed to two reasons: First, the Trojan policy $\pi_{fail}$ learned by the Trojan agent through our considered adaptive attack performs less effective compared with that for BackdooRL, which can be revealed by its accumulated reward is larger according to Line.6 in \cref{algorithm:mda}. Secondly,  adaptive BackdooRL would make the Trojan policy $\pi_{fail}$ fail slower compared with BackdooRL. As discussed in BackdooRL~\cite{backdoorrl}, the neural network for Trojan agents(\eg, LSTM~\cite{lstm}) has limited memory to remember the Trojan policy, therefore BackdooRL proposes to make $\pi_{fail}$ fail as quickly as possible. However, the adaptive BackdooRL may make the entire Trojan agent hard to imitate the Trojan policy.

Last but not least, we also summarize the failing speed for the adaptive BackdooRL and BackdooRL in \cref{fig:adaptive_2}. We can see that adaptive BackdooRL would require the Trojan agent to take significantly more steps for failing compared with BackdooRL. In the real world, failing speed would also affect the stealth of Trojan attacks against reinforcement learning. This is because if a Trojan agent takes more steps to fail, it would be more likely to be observed and taken controlled by the human beings. From this perspective, adaptive BackdooRL would behave less stealth in the real world application compared with BackdooRL.

Considering the attack efficacy and failing speed for the adaptive BackdooRL, we think the attacker may not have much incentive to conduct the existing Trojan attack(\ie, BackdooRL) and its variants against PolicyCleanse. However, there may be other advanced Trojan attacks to bypass PolicyCleanse while preserving the attack efficacy, which can be explored in the future.

{\small
\bibliographystyle{ieee_fullname}
\bibliography{example_paper.bib}
}

\end{document}